\documentclass[preprint,12pt]{elsarticle}

\usepackage{amsmath,amsfonts}
\usepackage{algorithmic}
\usepackage{algorithm}
\usepackage{array}
\usepackage{trace}
\usepackage[caption=false,font=normalsize,labelfont=sf,textfont=sf]{subfig}
\usepackage{textcomp}
\usepackage{stfloats}
\usepackage{url}
\usepackage{verbatim}
\usepackage{graphicx}
\usepackage{xcolor}
\usepackage{cleveref}

\DeclareMathOperator*{\E}{\mathbb{E}}
\def\grad{\nabla}

\journal{CNSNS}

\begin{document}

\begin{frontmatter}



\title{An effective control of large systems of active particles: An application to evacuation problem}


\author[airi]{Albina Klepach}
\author[sk]{Egor E. Nuzhin}
\author[sk]{Alexey A. Tsukanov}
\author[sk,uol]{Nikolay V. Brilliantov\corref{cor1}}

\cortext[cor1]{Corresponding author.}

\affiliation[airi]{
    organization={AIRI}, 
    city={Moscow}, 
    postcode={121170}, 
    country={Russia},
    note={Work done while at the Applied AI Center, Skolkovo Institute of Science and Technology}
}

\affiliation[sk]{
    organization={Artificial Intelligence Center, Skolkovo Institute of Science and Technology}, 
    addressline={Bolshoy Boulevard, 30, bld.1}, 
    city={Moscow}, 
    postcode={121205}, 
    country={Russia}
}

\affiliation[uol]{
    organization={Department of Mathematics, University of Leicester}, 
    addressline={University Road}, 
    city={Leicester}, 
    postcode={LE1 7RH}, 
    country={United Kingdom}
}
\ead{nikolaigran@gmail.com}

\begin{abstract}
Manipulation of large systems of active particles is a serious challenge across diverse domains, including crowd management, control of robotic swarms, and coordinated material transport. The development of advanced control strategies for complex scenarios is hindered, however,  by the lack of scalability and robustness of the existing methods, in particular, due to the need of an individual control for each agent.
One possible solution involves controlling a system through a leader or a group of leaders, which other agents tend to follow. Using such an approach we develop an effective control strategy for a leader, combining reinforcement learning (RL) with artificial  forces acting on the system. To describe the guidance of active particles by a leader we 
introduce the generalized Vicsek model. 
This novel method is then applied to the problem of 
the effective evacuation by a robot-rescuer (leader) of large groups of people from hazardous places. We demonstrate, that while a straightforward application of RL yields suboptimal results, even for advanced 
architectures, our approach provides a robust and efficient evacuation strategy.     
The source code supporting this study is publicly available at: \url{https://github.com/cinemere/evacuation}.
\end{abstract}

\begin{keyword}


Control theory  \sep Reinforcement learning \sep   Active matter  \sep Artificial forces 
\end{keyword}

\end{frontmatter}



\section{Introduction}

The manipulation of large systems  of active particles, especially controlling their collective behavior, has become a fundamental problem in recent decades, initiated by emerging new areas of application. 
The examples, across diverse domains, 
include crowd management \citep{helbing1995social, Warren2018CollectiveMI}, controlling of robotic swarms  
\citep{swarm_robots1}, coordinated material transport \citep{Uspal2018TheoryOL}, etc.  The concept of active matter  \citep{VicsekINITIAL} offers an appropriate framework for modeling such ensembles, 
comprised of agents that consume energy to move, interact with their environment, and adjust their direction of motion subject to external signals \citep{doi:10.1177/10597123231202593,OLCAY20191,BAR2024101701,Egor2021}. 
Synthetic microswimmers,  swarming robots, colonies of bacteria, fish schools, flocks of birds, groups of humans   \citep{RevModPhys.88.045006,Egor2021} -- all these systems of living and non-living agents represent an active matter.  

The primary goals of controlling active particle systems include navigating them in complex environments and supporting synchronized collective behavior \citep{microsw1,microsw2,microsw3}. This can be addressed by classical optimal control theory, which requires a complete knowledge of the environment and dynamics (similar to Zermelo’s navigation problem \citep{Zermelo}), or by reinforcement learning (RL), where an agent learns strategies through trial-and-error \citep{RL,RL2}. The latter is particularly applicable when an agent has only partial, local information or when the impact of noise is significant. A key limitation of such control problems 
is that their solution relies on the individual manipulation of each participating particle. For ensembles comprising hundreds of agents a  straightforward application of RL, with individual-level control, becomes computationally intractable.

To overcome this limitation, a common strategy involves guiding a group through a leader or multiple leaders, when  active particles just  follow the leader(s)  \citep{Seyboth2014CooperativeCO,Nowzari2017EventtriggeredCA,Hu2021ADC,Long2019ACR}. This leader-follower paradigm significantly simplifies the control problem by focusing strategies solely on the leader(s). Nevertheless, complexities persist, particularly when managing multiple groups of different  size and location, or when groups cannot be manipulated simultaneously or possess non-coinciding goals \citep{Bernardi_2021,Albi2015InvisibleCO}. Among 
important examples  of such challenging problems is an effective guidance of large groups of people, especially -- their evacuation from  hazardous places. Here a leader (rescuer) 
has the goal to evacuate all people 
guiding them to the exit(s) in shortest time.
 Given the inherent randomness and non-negligible noise within these systems, reinforcement learning (RL) seems to be the most 
 suitable approach. However, a straightforward application of standard RL 
 is neither computationally efficient nor effective in achieving  the ultimate goal.

To this end, we propose an application of  auxiliary (artificial) 
``pseudo-gravitational'' forces acting on the system, as a part of the environment; this  helps the leader to find the most effective guidance strategy. 
Such   a combination of RL with artificial forces \citep{KALA2024465} results in a very  effective method to control large ensembles of active particles.
To implement our new method for the evacuation problem,  we utilize the generalized Vicsek model. This model takes into account 
not only interactions between active particles
(as in the conventional Vicsek model \citep{VicsekINITIAL}), but also between particles and the leader.

The rest of the paper is organized as follows. In the next Sec. II we formulate the problem of an effective evacuation from a hazardous place by an informed rescuer-leader. In Sec. III we describe the methods for training optimal evacuation policy and propose an approach based on ``pseudo-gravitational'' forces, viz the new method of RL with artificial fields.  In Sec. IV we report the results of our numerical experiments. Finally, in Sec. V, we summarize our findings. 

\section{Effective evacuation problem} \label{evacuation_problem}

\subsection{General statement of the evacuation problem } \label{evacuation_overview}
The problem of effective  evacuation, especially in the case of panicking crowd is extremely important. Poor building planning or lack of awareness of escape routes in hazardous events (natural disasters, terrorist attacks, etc.) can lead to fatalities for many reasons, including stampedes \citep{Borve,Pradeepkumar,Peacock,Liu}. 

In planning a safe evacuation strategy,  the following  problems are to be addressed: What is the fastest way to evacuate people from the buildings that have already been built? How can we erect safe evacuation structures that allow quick escapes in  emergency? Furthermore, it is  important to have a representative quantity to assess an evacuation strategy.

Generally, evacuation studies employ several main methods \citep{fang2022evacuation}: indoor experiments with real people \citep{NATAPOV2022105483}, numerical simulations (often incorporating machine learning for optimization) \citep{Borve,7332972,8989970,9442521}, and hybrid approaches combining both \citep{7373229,wang2019improved,7152824}. The goal of such an analysis is to reveal some rules of crowd behavior, translate them into mathematical models, and reconstruct the crowd dynamics in numerical experiments. Machine learning can be used to optimize the realism or efficiency of evacuation.

Among the first studies of evacuation was Helbing's work \citep{helbing1995social}, where the social force model (SFM) has been pioneered. Based on visual patterns of people's behavior, subject to a combination of physical interactions and the action of the social norm of keeping apart from unknown persons, the authors proposed an appropriate mathematical model. SFM was then successfully applied to describe panicking crowds \citep{9442521}. This model is very general, and may be easily adopted, by respective modifications, to various special cases. SFM-like models are able to reproduce various self-organizing phenomena, such as lane formation, clogging at exits \citep{24a,Helbing2011,Bodrova2024} customer dynamics \citep{tsukanov2021}, etc. 

The main limitation of such models is the computational complexity, associated with the complete implementation of the SFM for a large system of agents. Therefore the lattice-automata approach, was widely used, with the set of discrete actions and states. Here the pedestrians are individual "particles", with an appropriate speed and size; their movement is governed by spatial transition probabilities. This allows encoding of simple rules of behavior in numerical simulations, which is realized in a number of available software ( Legion \citep{Legion}, EXITUS \citep{7332972}, Simulex \citep{THOMPSON1995149}, EXIT89 \citep{Fahy1994Exit8E},
 PEDFLOW \citep{PEDFLOW}, etc.) 

Further development of evacuation models has continued for the last two decades, and numerous new models have been suggested. These include models that allow evacuation testing of a building using the respective simulations at the planning stage \citep{SHIN2019545}.  Similarly, there are several studies, where various evacuation strategies can be tested for buildings already built \citep{Kuligowski2010,Kisko1985EVACNETAC,ABIR2022103083,zhang2021crowd,pugh2021deep}.

Planning the safest escape is a complicated problem, which includes simulations of human behavior in emergencies. 
It requires the appropriate planning of the best routes that people should follow during evacuation from dangerous places. 
The overall efficiency of the evacuation strategy is to be investigated for a particular building layout and sources of danger. To find the optimal routes, one needs to deal with a large number of variables, associated with the locations of individuals and geometry of the evacuation space.  

Various methods and models have been proposed to solve this problem. For example, a Markov Decision Process with intermediate decision-points has been employed \citep{pugh2021deep,tsukanov2021}, incorporating the Social Force Model (SFM) \citep{helbing1995social} within a discrete state-action space. The problem has also been framed as multi-agent learning \citep{wang2019improved}, treating groups of evacuating people as agents. To realistically describe inter-person interactions, SFM modifications have been explored, such as attributing group formation to attractive interactions. A two-level hierarchical approach, combining macro-level path programming with micro-level pedestrian-obstacle and inter-pedestrian interactions, was used in 
\citep{zhang2021crowd}.

In the present study, we analyze the evacuation by an informed rescuer-leader, which guides the pedestrians towards the exit. 
We explore the conventional Viscek model \citep{VicsekINITIAL} in the context of  pedestrian motion dynamics and propose the "generalized Viscek model" to account for the leader. Futher, we apply the method of reinforcement learning (RL) \citep{RL} to train the leader's policy. The leader is supposed to be a robot which acts according to an optimal strategy, found in a learning process. The idea of a robot-assistant evacuation seems to be rather fruitful, see e.g. a review \citep{Roboteva}. Further, we develop this approach and describe the problem setting. 

\subsection{Generalized Vicsek model}
Here we introduce a generalized Vicsek model,  which describes motion of people, guided by a leader.  Generally, a motion of a person surrounded by a crowd may be decomposed into two parts -- the one  depending only on the  person itself,  and the other, imposed by the crowd. Often this results in self-organized patterns. One of the simplest models quantifying self-organization of moving agents,  is the celebrated Vicsek model \citep{VicsekINITIAL}. In spite of its simplicity, it demonstrates a rich  behavior, with a large potential for further extensions, see e.g. \citep{PhysRevLett.75.4326,Gr_goire_2003,labasken,collective_turns}. The model is based on a few simple rules and takes into account random fluctuations. In the original version 
{\citep{VicsekINITIAL}}, the  main rule states, that each time step a moving agent adopts its direction of motion, to the average direction of all neighbours located within the distance \(r\) from the agent. The absolute velocities of the agents are kept constant and the system is subjected to a white noise. Mathematically, the Vicsek model is formulated for a two-dimensional motion as following: 
\begin{equation}
    \begin{aligned}
        \begin{cases}
            \vec{r_i}\texttt{(t+}\Delta \texttt{t) = }\vec{r_i}\texttt{(t) + }v \vec{e_i}\texttt{(t)}\Delta \texttt{t} \\
            \theta_i\texttt{(t+}\Delta \texttt{t) = } \langle \ \theta_k\texttt{(t)} \ \rangle_{ |\vec{r}_i - \vec{r}_k| < r} \texttt{ + } \zeta_i\texttt{(t)}  
        \end{cases}
        i \texttt{=} \overline{1N},
    \end{aligned}
\label{VM_equations}
\end{equation}
where 
\(\vec{r_i} (t)\) and \(\vec{r_i} (t+\Delta t)\) are respectively the coordinate of \(i\)-th particle (in the ensemble of \(N\) particles) at time \(t\) and \(t+\Delta t\). Each particle has a velocity \(\vec{v}_i(t)= v \vec{e}_i(t)\) at time \(t\), which absolute value is \(v\) and the direction is determined by the unit vector 
\( \vec{e_i}\texttt{(t) = }\left(\cos\theta_i\texttt{(t)}, \sin\theta_i\texttt{(t)}  \right)^{T} \), with the  azimuthal angle \(\theta_i\). The  radius of neighbour influence is \(r\), and \(\zeta_i(t)\)  is a random value, corresponding to the noise. In the original Vicsek model it is a random number with a uniform probability from the interval \([-\eta/2, \eta/2]\), where \(\eta\) is the noise magnitude.

The conventional Vicsek model reflects an important feature of a crowd  -- the so-called "herd mentality" \citep{rast1963effects}.  It reflects the empirical observation that the individuals may be influenced by their peers and adopt their behavior. {The main difference between the current and standard Vicsek model is the introduction of a leader with an independent action policy. While in the original Vicsek model all individuals are peers, the motion of an individual in our model is affected by both -- by the motion of the crowd and that of a particular individual, called leader}. {The present model assumes a single leader, although it may be straightforwardly generalized for multiple leaders.}  The leader is characterized by a stronger influence on the crowd, as compared to a common individual. Namely, an individual will preferably follow the direction of a leader and not of the neighbors. {This extension is essential in the evacuation problem, where the leader is responsible for guiding individuals to the exit.} To take into account an impact of a leader on a  motion of a person, we generalize accordingly Vicsek model. We assume that the leader possesses its influence zone --  a circle of  radius \(r_{\rm L}\), and its influence is characterized by the coefficient \( q \in [0, 1] \) -- the  \emph{enslaving degree},  which quantifies the relative influence of the leader and neighbors. Hence we formulate the generalized Vicsek model as following\footnote{{Note that Eq. \eqref{MVM_update_rule} is written in the original notations of Vicsek model, so that the expression $q\,\theta_{l}(t) + (1-q)\,\theta_{\rm v,i}(t)$ denotes the  weighted average direction of the two directions 
with angles $\theta_{l}(t)$ and  $\theta_{\rm v,i}(t)$ and corresponding weights $q$ and  $1-q$. 
This averaging is performed according to the procedure described in \ref{sec:average_angle} [see Eqs. \eqref{anglesystem}, and \eqref{eq:p_xy_weigted}].}}.
\begin{figure}[ht] 
\centering
\includegraphics[width=0.7\textwidth]{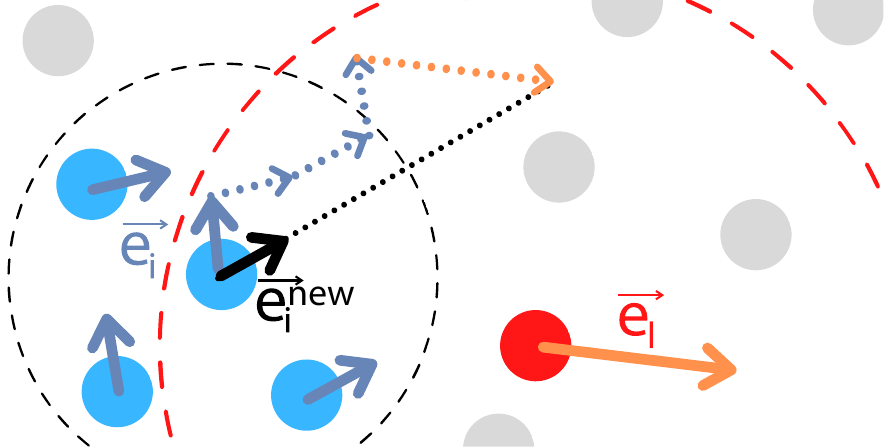}
\caption{\label{VM_scheme} Sketch, illustrating the generalized Vicsek model. The small filled circles represent the individuals (grey and blue) and the leader (red). The smaller and larger dashed circles denote respectively, the Vicsek influence zone and the influence zone of the leader. The  blue filled circles,  residing  within the interaction circle of the Vicsek radius, are the neighboring individuals  that interact with the agent in the center of the smaller dashed circle. The grey filled circles represent distant persons, that do not interact with this individual.  The arrows indicate the velocities of the agents. Blue arrows are the velocities at the previous time step, the orange  arrow is the velocity of the leader and the black arrow is the new velocity of the individual in the center of dashed circle,  after the application of the generalized  Vicsek rule.}
\end{figure} 
\begin{equation}
    \begin{aligned}
        \begin{cases}
           \vec{r_i}\texttt{(t+}\Delta \texttt{t) \!\!= \!\!}\vec{r_i}\texttt{(t) + }v \vec{e_i}\texttt{(t)}\Delta \texttt{t} \\
            
            \theta_i\texttt{(t+}\Delta \texttt{t) \!\!= \!\!} 
            \begin{cases}
                \begin{aligned}
                    q \theta_{{l}}\texttt{(t)} \!+ \!(1 - q)\theta_{\rm v,i}\texttt{(t)} 
                    & \quad \!
                    \text{if \( |\vec{r}_i - \vec{r}_{l}| < r_\texttt{L}\)}\\
                    \theta_{\rm v,i}\texttt{(t)} 
                    & \quad
                    \text{else}
                \end{aligned}         \end{cases}
        \end{cases}
    \end{aligned} 
\label{MVM_update_rule}
\end{equation}
where  \(\vec{r}_{l}\) is the coordinate of the leader, which moves with the velocity \(\vec{v}_{l}= v \vec{e}_{ l}\), with \(\theta_l(t)\) being the azimuthal angle that determines the direction of motion of the leader, \(\vec{e}_l(t)\) and \(\theta_{\rm v,i}(t)\) defines the direction of motion of the individual, subject  to the Vicsek's rule only, that is, 
\begin{equation*}
    \theta_{\rm v,i} = \langle \ \theta_k\texttt{(t)} \ \rangle_{ |\vec{r}_i - \vec{r}_k| < r} + \zeta_i\texttt{(t)} 
\end{equation*}
Hence the  additional control parameters of the generalized Vicsek model, as  compared with the Vicsek model are the enslaving degree \(q\) and the influence radius of the leader \(r_{\rm L}\). The generalized Vicsek model is illustrated in \ref{VM_scheme}.

\subsection{Evacuation simulation}

The evacuation problem is studied within the framework of evacuation from a dimly lit (dark, smoky, {etc.}) room with a single exit. The movement of escaping individuals and the leader-rescuer is subject to the generalized Vicsek model. 

\begin{figure}[ht]
\centering
\includegraphics[width=0.7\textwidth]{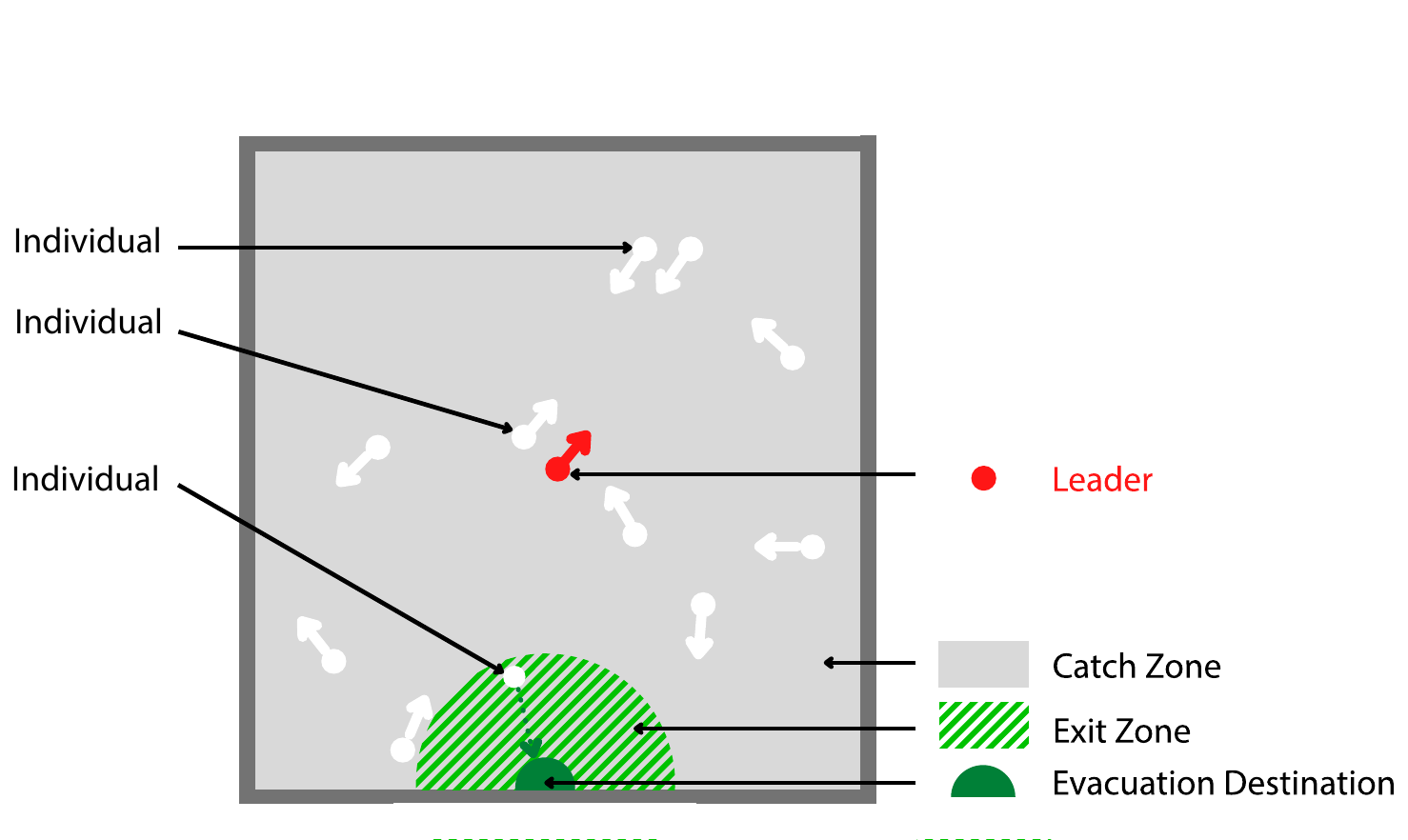}
\caption{\label{env_scheme} Sketch of the evacuation environment. The white arrows represent individuals; the directions of the arrows indicate the direction of their motion. The red arrow depicts the position and motion direction of the leader.}
\end{figure}

\paragraph{Environment dynamics, observations and actions} 
The space of the room is divided into two zones, suggesting different strategies for the leader: the catch zone and the exit zone, which are marked in \Cref{env_scheme}. The simulation begins by placing the leader in the center of the room and randomly distributing people. The leader's task is to guide to the exit all the people,  as fast as possible.
The behavior of the leader and individuals is the following. 
In the first,  \emph{catch} zone, see \Cref{env_scheme}, the leader's task is first to collect the individuals, and then guide them to the exit. The leader's action policy is defined by an artificial neural network (ANN), which prescribes the direction of the leader motion. {The ANN serves as a parameterized function whose weights are trained using RL to achieve an effective evacuation}.  The individuals move initially randomly,  but obey the laws of the generalized Vicsek model, Eq. \eqref{MVM_update_rule}. That is, {they have tendency to follow the leader} 
if they are in its area of influence; they can  be also "captured" by other individuals and move in the same direction as the group.

In the \emph{exit} zone, individuals do not interact with the leader, but move exactly towards the exit, obeying the following relations:
\begin{equation}
\begin{cases}
    \vec{r_i}(t+\Delta t) = \vec{r_i}(t) + \vec{v_i}(t)\Delta t \\
    \vec{v_i}(t+\Delta t) = v\ \vec{e}_{i \, {\rm exit}}(t)  
\end{cases}
i = \overline{1N},
\end{equation}
where \(v\) is {and absolute velocity of the generalized Vicsek model} the and 
\begin{equation}
    \vec{e}_{i\, {\rm exit}}(t) = \frac{\vec{r}_{\rm exit} - \vec{r}_i(t)}{|\vec{r}_{\rm exit} - \vec{r}_i(t)|}\\
\end{equation}
is the unit vector for \(i\)-th particle to the exit destination.
At the end, having reached the evacuation destination -- the small circle with an exit point at the center, \Cref{env_scheme}, an individual is considered saved. For clear simulations and training, the environment has been  implemented  with the interface of OpenAI Gym environment\footnote{Source is available on GitHub: \url{https://github.com/cinemere/evacuation} with the interface of OpenAI Gym environment (\url{https://gymnasium.farama.org})\citep{brockman2016openai}.}.

The leader's action is defined as a two dimensional vector \(\vec{a} \in \mathbb{R}^2\) pointing the direction of the leader motion. For simplicity of the technical implementation,  we do not apply initially, constraints on its length. Later, we normalize it, to obtain the unit vector \(\vec{e}_{ l}= \vec{a}/|\vec{a}|\), used in the generalized Vicsek model. Chosen an action, the leader updates its position, \(\vec{r_l}\), following the dynamics
\begin{equation}
            \vec{r_l}\texttt{(t+}\Delta \texttt{t) = }\vec{r_l}\texttt{(t) + }v \vec{e}_{ l}\texttt{(t)}\Delta \texttt{t} ,
\label{leader_equations}
\end{equation}
with {\(\vec{e}_{ l}(t) = \vec{a}(t)/|\vec{a}(t)|\)}. 

The observation space, that is, what the leader sees, comprises the coordinates of all the  individuals {in the system }
\(\{\vec{r}_{i}\}\), and coordinates of the leader, \(\vec{r}_{l}\), itself: 

\begin{equation}
\label{s}
    {s = \{\vec{r}_{l},\vec{r}_{1},...,\vec{r}_{N}\}, }
\end{equation}
{where \(N\) is the total number of the individuals (in the simulations this number is kept constant).}

{At the beginning of each simulation episode, the rescuer-robot is placed at the center of the room. The individuals are initialized with random positions sampled uniformly within the room area, and their initial velocity directions are assigned randomly from a uniform distribution over angles. This initialization procedure is used consistently across all experiments reported in the paper.}

\paragraph{Reward} \label{subsection:Reward}

The goal of RL is to find an optimal policy of the leader to act, that is, to choose the direction of motion, \(\vec{e}_{l}=\vec{a}/|\vec{a}|\), for each given state of the environment, \(s\). This is described by the 
probability density function \(\pi(a|s)\), which specifies the leader's action probability to choose an action \(a\) being in the state \(s\).

{\begin{algorithm}[ht]
    \small
    \begin{algorithmic}[1]
    \FOR{\textsc{each step} \texttt{t} \textsc{of the simulation} \COMMENT{\(\texttt{t} = 1,2,\dots,\texttt{t}_\texttt{max}\)}}
            \STATE \(\tau \gets 1 - \dfrac{\texttt{t}}{\texttt{t}_\texttt{max}}\)
            \STATE \(r \gets \bigl(w_A + w_B\,\tau \bigr)\, N_\texttt{exiting}^\texttt{new}\texttt{(t)}\) \COMMENT{positive reinforcement for each individual entering \emph{the Exit Zone}} 
            \STATE \(R_t \gets  r - p\) \COMMENT{complete reward taking into account  positive reinforcement and time step penalty}           
            \STATE \textsc{save:} \(R_t\)
    \ENDFOR
    \end{algorithmic}
    \caption{{Reward Computation}}
    \label{alg:reward}
\end{algorithm}}

The reward is used to train the leader in the evacuation environment, to achieve the best action policy. Here the main goal is to reduce the total evacuation time of the individuals, as much as possible. To achieve this goal an appropriate reward is to be tailored. The straightforward application of RL implies that a reward is given only after achieving the goal. This however  results in a sparse rewarding, which is computationally very challenging, therefore we apply an approach of intermediate reward for low-level goals, {similar to} Ref. \citep{dorigo1998}, and give positive reinforcement for each individual entering the exit zone.
{In Algorithm~\ref{alg:reward} we describe the procedure of reward computation. 
Here, \(R_t\) denotes the reward at time step \(\texttt{t}\); 
\(N_\texttt{exiting}^\texttt{new}(t)\) is the number of individuals who just entered the Exit Zone at time step \(\texttt{t}\); 
\(\texttt{t}_\texttt{max}\) is the maximum episode length; 
\(w_A=15\) and \(w_B=10\) are empirical parameters influencing the amount of positive reinforcement; 
and \(p=1\) is the step penalty for the time spent.}

\section{Methods}
Generally, our approach is based on the RL method, with an appropriate modification to increase evacuation efficiency, and the model of environment, where a trainable agent (robot-rescuer) is acting. Here we discuss the respective modifications of the RL; its standard part is briefly sketched in the \ref{rl_overview}. {In our implementation the policy \(\pi(a|s)\) is a parameterized function represented by a neural network. The network outputs the mean and variance of a Gaussian distribution over actions, conditioned on the observed state. This distribution is then used both for sampling actions during training and for evaluating action probabilities.}The environment, in its turn,  comprises a number of individuals to be rescued, properties of the space, where people are located and position of the robot itself. The individuals interact with each other and the robot. 

\subsection{``Pseudo-Gravitational'' potentials }
\label{sec:set_symmety}

The idea to exploit potential-field methods to  guide an agent was very fruitful in robotics, see e.g. \citep{Lynch,Khatib,Koren,Devlin}.  Here the goals and obstacles are treated as artificial "fields" that exert virtual forces on a robot, helping in this way its navigation. In other words, a goal creates an "attractive" basin, while obstacles create "repulsive" hills, and the agent moves under the resulting force $-\nabla U$. Because only local gradients are needed, the method yields fast, reactive navigation. Analogous ideas appear in reinforcement learning -- by adding a “potential” term to the reward, one shapes a value landscape that nudges the learner in desirable directions while preserving the optimal policy. Hence artificial potential fields give an elegant, physically motivated way to steer agents. Inspired by these ideas we propose artificial force fields, which are also gradients of the respective artificial potentials. Since the potentials are expressed in terms of the inverse power of the according distances, we call them ``pseudo-gravitational''.  Note, however, that the nature of our potentials differ from this of standard artificial  potentials, since they include the number of active particles in a group; below we present their  detailed definition.

The environmental conditions, observed by a leader, are  described by coordinates of individuals inside the building.  
Hence, the action policy of the leader is to be formulated in terms of these coordinates.

There are different ways of defining the policy. For instance, {the policy can be defined through a feed-forward neural network that straightforwardly processes the full set of individual positions as its input (note, that in our setting the state space is continuous and infinite). Such an approach may be, however, prohibitively expensive, especially with respect to RL optimization.}\footnote{{This approach either assumes that the high-dimensional state can be directly compressed into a lower-dimensional latent representation, or requires scaling the network width proportionally to the input dimension. Both options make training inefficient or computationally demanding.}}
Alternatively, since all individuals are equivalent, it would be more beneficial to deal  with some {appropriate } collective characteristics. This suggests an implementation of  the Deep Sets encoding \citep{DeepSets}. Here we deploy this approach with a specific encoding possessing  a physical interpretation, which may be  called ``pseudo-gravitational''  encoding. {Note, that such a reduction is particularly important for large ensembles. Indeed, a straightforward RL approach would require processing of the full state of all individuals; this becomes intractable for sufficiently large ensembles.}

Let us remind, that the goal of the rescuer-leader is to guide as much as possible people to the exit of a building, when the individuals are not aware about its location, due to the darkness, smoke, etc. At the same time the rescuer-leader ("leader",  below, for brevity) possesses a complete information about the location of all the individuals and the exit. The leader needs to approach an individual, or a group of individuals to a "catching" distance -- the distance where they start to follow the leader, and guide  them to the exit. A straightforward approach to learn the efficient policy, is to directly use the whole information, about all individuals and the leader location. Such an approach, which we call  "no encoding" may experience learning difficulties, especially for a large number of individuals. Hence, it is reasonable to look for a cumulative characteristics  for the groups of people and their locations.

Obviously,  more people would be saved if the leader guided to the exit a large group of people, than a small group.  On the other hand, if a small group is located close to the exit, it seems reasonable to  guide it first to the exit, and only  then,  move back to a larger group, residing  further from the exit. The ``pseudo-gravitational'' rules provide the mathematical background for such a common sense based strategy. 

The idea is to associate all positions of individuals,  with respect to the leader, with a certain attractive, gravity-like phantom potential ("catch" potential). The position of the leader, with a number of followers, with respect to the exit, is to be associated with another phantom potential { -- the  "exit"  potential}. Then, the decision is to be done, based on the phantom forces -- "pseudo-gravity {forces}", which are  the gradients of {such potentials}.   
 
This way we help the leader to decide, whether to move towards a remote group of people, or to the exit, that is, we help to find the trade-off between collecting new individuals, or escorting already collected ones to the exit. Hence we write for both pseudo-gravitational potentials:
\begin{align}
\label{potential}
    U_{\rm catch}(\vec{r}) &= -\sum \limits_{\rm i \in  free} |\vec{r} - \vec{r}_i|^{-\alpha},\\
    U_{\rm exit}(\vec{r}) &= - {|\vec{r}_{\rm exit} - \vec{r}|}^{-\alpha} \cdot N_{\rm caught}.
\end{align}
Here \(N_{\rm caught}\) is the number of individuals collected by the leader (that is, the individuals which currently follow the leader) and \(\vec{r}_{\rm exit}\) gives the location of the exit. The exponent \(\alpha \ge 0\) describes the dependence of the potential on the corresponding distances. 

The gradients of the potentials are associated with the effective "forces", which are utilized in the decision making; these forces form the  "environment {perception}" for the leader in the RL.

\begin{align}
\label{gravity}
    F_{\rm catch}(\vec{r}) &= -\grad_{\vec{r}} \; U_{\rm catch}(\vec{r})\\
    F_{\rm exit}(\vec{r}) &= -\grad_{\vec{r}} \; U_{\rm exit}(\vec{r}).
\label{gravity_exit}
\end{align}

\subsection{Training details} \label{subsec:methods}

In the present work, we compare three methods: \emph{feed-forward}, \emph{transformer}, and a novel approach based on \emph{pseudo-gravitational} encoding. The primary distinction among these methods lies in their neural network architectures. {It is important to note that the pseudo-gravitational encoding is not a standalone neural architecture, but rather a state-representation scheme that acts as the first layer of the network. The encoded forces are then passed to the same feed-forward network architecture as in the baseline model, with the only difference being the input representation.} To train the action policy, we utilized the Robust Policy Optimization (RPO) \citep{rahman2022robust} implementation from the CleanRL \citep{huang2022cleanrl} package. The complete set of RPO agent hyperparameters is summarized in \Cref{num:table:RPOAgent}, while additional details on the RL algorithm are presented in \ref{rl_overview}.

\paragraph{Feed forward architecture} In the straightforward application of RPO, the leader's (training agent's) observation 
{ i.e. the input vector is represented by the leader’s coordinates, relative coordinates of the exit (relative to the leader) and the relative coordinates of all $N$ individuals. 
For evacuated individuals, we replace their  positions by the exit vector 
relative to the leader, $\vec{r}_{\text{exit}} - \vec{r}_{l}$; this ensures that the original indexing remains unchanged.
Thus, the input dimension is always $2N+4$.} This forms the input to the policy network, {\(\pi (a\ |\ s_{FF})\), where \(s_{FF} = (\vec r_{l},\vec r_{exit} - \vec r_{l},\vec r_{1}-\vec r_{l}, \vec r_{2} - \vec r_{l}, \dots, \vec r_{N} - \vec r_{l})\)}, which then predicts the leader's action: a two-dimensional motion vector \(\vec{a}=(a_x,a_y)\). The network parameters are provided in \Cref{num:table:Network}.

\paragraph{Transformer architecture} To better capture interrelations between particles, we incorporated a learnable attention mechanism into {ANN} 
\citep{vaswani2017attention}. Here, information concerning the exit, leader, and pedestrians is concatenated into a single token-vector. {In general, a token in a transformer is the smallest unit of information that the model treats as an independent entity. Each token has its own vector representation and participates in the self-attention mechanism, which learns relations between tokens. The choice of tokens is problem-dependent: in NLP they are words, in the computer vision they are image patches. In our case, it is natural to define tokens as pedestrians, the leader, and the exit, since these are the entities whose interrelations drive the evacuation dynamics. 
Formally, we represent each token as a tuple $T=( \vec r,i)$, where $r$ is the
position vector of the entity (relative to the leader, unless otherwise specified),
and $i$ is a one-hot indicator of its type or zone. That is, we define:
\begin{itemize}
    \item $T_j = ( \vec r_j- \vec r_l,\, i_j)$ for $j=1...N$, where $i_j$ is one-hot encoding whether pedestrian $j$ is in the Vicsek zone, the catch zone, the exit zone, or already evacuated.
    \item $T_{exit} = ( \vec r_{exit}- \vec r_l,\, i_{exit})$, where $i_{exit}$ is the one-hot encoding corresponding to a pedestrian in the exit zone;
    \item $T_l = ( \vec r_l,\, i_l)$, where $i_l$ is always the zero vector, since the leader does not belong to any pedestrian zone.
\end{itemize}}
Each token thus provides a compact
description of one entity, and the self-attention mechanism allows the model
to capture interactions between these entities directly.

Attention is estimated by adding BERT-style \citep{kenton2019bert} transformer block(s), which transforms {representation} of each token: \( T \rightarrow T'\) allowing the model to focus on relevant input data and enhancing computational efficiency. The feed-forward policy then takes these transformed tokens as input: 
{\[\pi (a\ |\ {T'}_{l}, {T'}_{exit}, {T'}_{1}, {T'}_{2}, \dots, {T'}_{N}).\]} 
{The transformer blocks as well as the feed-forward decoder were jointly trained via reinforcement learning.}
The exact parameters for the transformer encoding are detailed in \Cref{num:table:TransformerBlock}, and those for the subsequent feed-forward network are in \Cref{num:table:Network}.

\paragraph{Pseudo-gravitational encoding architecture} The neural network with pseudo-gravitational encoding includes an additional initial layer without learnable parameters. The role of this layer is to "condense" the complete information of the environmental observation into cumulative characteristics. These characteristics are defined such that their dimension, or the number of output variables, does not depend on the number of individuals in the environment. For this purpose, we employ pseudo-gravitational forces as these cumulative characteristics. Specifically, instead of directly feeding the raw environmental observation ({relative} coordinates of all individuals {and exit} 
position) to the main neural network, this initial layer processes to compute the gradients of the pseudo-potential—namely, the aggregate "catch force" and "exit force" (defined by \Cref{gravity,gravity_exit}) {supplemented with the leader's own position to understand its spatial location}. These pre-calculated forces then serve as the primary, fixed-dimension input to the subsequent layers of the neural network (the policy \(\pi (a\ |{r_l},\ F_{catch}, F_{exit})\)), allowing the model to efficiently learn a policy based on these key aggregate features rather than individual particle positions. The exact algorithm for training a policy with Pseudo-Gravitational Encoding Architecture is presented at \Cref{alg:learninig}. The parameters for the feed-forward network are in \Cref{num:table:Network}.

\section{Experimental results}
Below we present simulation results for the solution of the evacuation problem by the new method. Numerical experiments have been performed to model evacuation of 60  individuals from a dimly lit room by a robot-rescuer (see also {\ref{apx:exp_res}} for the results for 15 individuals). The robot learns to save people in the most efficient way, based on its experience, with the use of reinforcement learning (RL). 

\begin{figure}[ht]
\centering
\subfloat[][No followers]{\includegraphics[width = 1.7in]{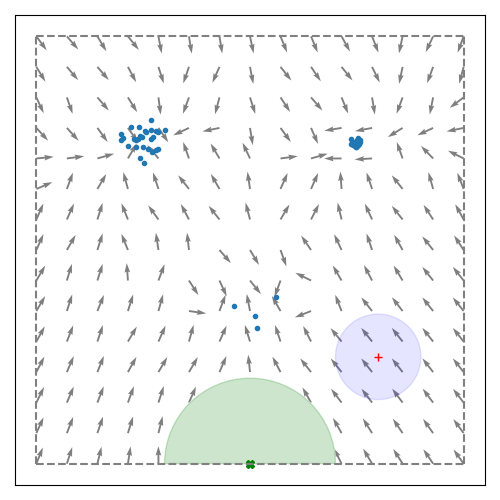}}
\hfil
\subfloat[][With followers]{\includegraphics[width = 1.7in]{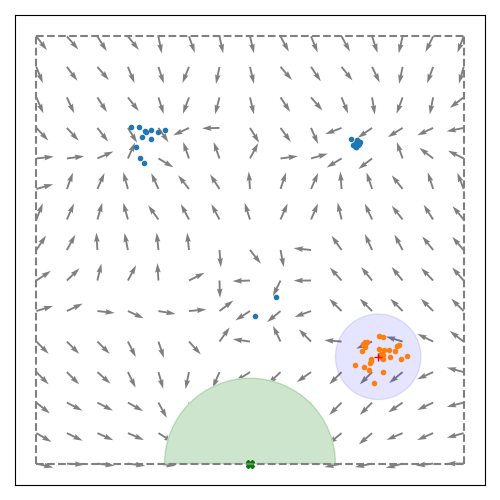}}
\label{policy}
\caption{Schematic representation of the optimal action policy.  
{The expected direction of the leader motion, in the corresponding point,} is indicated by grey arrows for the cases of:  (a) the leader has no followers and (b) half of the individuals are the followers. Blue circles correspond to "free" individuals, orange circles -- to the followers of the leader. The leader is designated by the red cross and its area of influence is shown by light blue.  Light green area is the exit zone, and green point is the evacuation destination.  }
\label{fig:policy}
\end{figure}

The optimal strategy depends on the number of people to be saved and the size and geometry of the room. Qualitatively, the trained policy is illustrated  in \Cref{fig:policy}. It represents the expected behavior of the robot-rescuer for different starting positions.  {Each arrow indicates the expected direction of motion of the robot-leader in the corresponding point.}
For instance, \Cref{fig:policy}a demonstrates, that the leader without followers tends to move towards individuals, and larger groups -- the top groups  attract the leader stronger than the small ones or single individuals, although located at lager distances. 
At the same time, the leader with a number of followers, as in \Cref{fig:policy}b, tends to move towards the exit, provided that the exit is close enough.

\begin{figure}[ht]
\centering
    \includegraphics[width =0.7\textwidth]{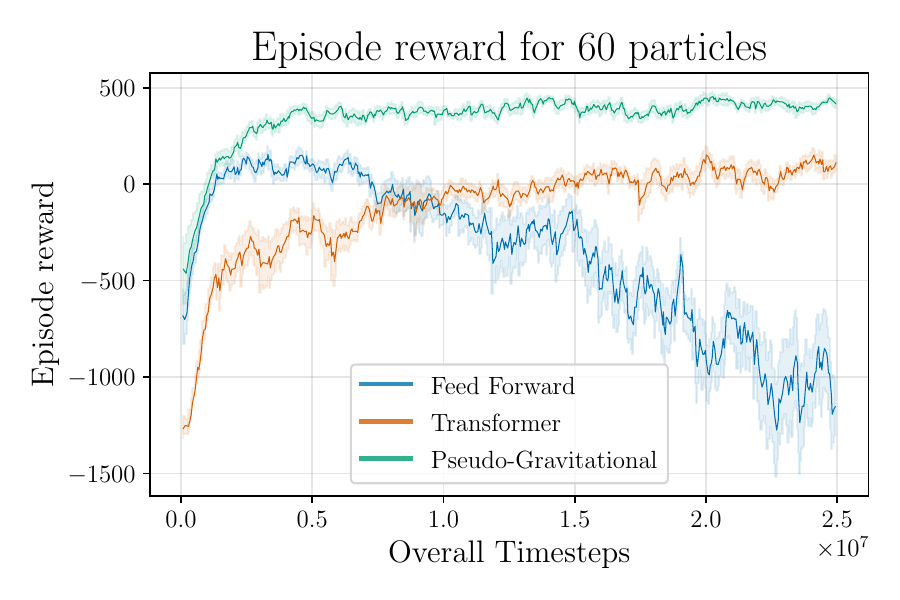}
    \caption{
     {The curves, representing the training process:} the average episode reward, { as a function of the simulation time step.} 
     The results are averaged over 10 training runs {with different environment random seeds}; the shaded areas depicts the standard error. The green curve corresponds to RL with the ``pseudo-gravitational'' encoding, orange - RL with Transformer architecture and  blue -- RL with Feed Forward architecture. The exponent of the ``pseudo-gravitational'' potential \(\alpha = 1\).}
    \label{diagram_training}
\end{figure}

In our simulations we observe that the rescuer-leader firstly learned to walk around the walls, then it learned to collect and escort individuals to the exit; finally it learned to do this as fast as possible. In \Cref{diagram_training} we show the learning kinetics -- the evolution with time of the average reward of a training episode\footnote{The averaging over time makes the results more interpretable and visually clear due to smoothing short-term fluctuations.} (we use here an exponential moving average \citep{brown1956exponential}). As it follows from the figure, the learning time is about \(6 \cdot 10^4\) time-steps -- after that training time the leader "knows" how to evacuate people in the most fast and effective way. In \Cref{diagram_training} we also show the learning curve for  RL with basic feedforward architecture (without ``pseudo-gravitational'' encoding), as well as for one of the most popular, state-of-the-art,  transformer-based architecture  \citep{vaswani2017attention}. As it is clearly seen from \Cref{diagram_training} (as well from its counterpart, \Cref{alpha_evac1} in the Appendix), the new algorithm with the ``pseudo-gravitational'' encoding substantially outperforms the algorithm with the transformer architecture. Moreover, it dramatically outperforms the  RL with basic feedforward architecture, which essentially fails to perform an effective evacuation.  
{In addition to higher evacuation efficiency, our method also exhibits faster and more stable convergence compared to the baselines. While transformer-based models require substantially more training steps and feed-forward models fail to converge, the pseudo-gravitational encoding consistently reaches effective strategies}
(see \Cref{alpha_evac1}). Such advantages of the new algorithm have been demonstrated for different scenarios -- for 60 and 15 individuals in a hazardous zone,  which is demonstrated in \Cref{diagram_training} and \Cref{alpha_evac1}. Interestingly, our experiments show that $\alpha=3$ is the optimal value for the parameter $\alpha$ of the ``pseudo-gravitational'' potential (see Appendix C and \Cref{alpha_evac}).

\begin{figure}[ht]
    \centering
    \includegraphics[width=0.7\textwidth]{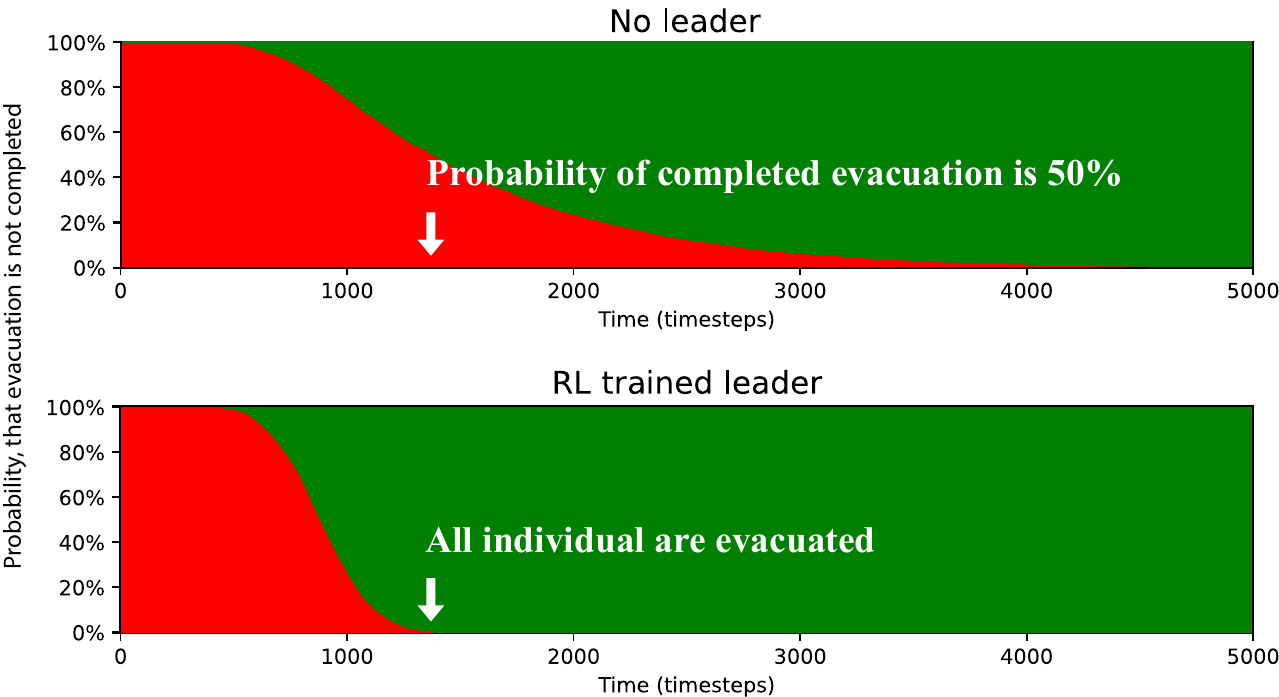}\\
    \caption{ {Illustration of the evacuation efficiency, assessed by the probability of evacuation of all individuals  before time $t$.} Upper panel: Self-evacuation without a  leader. Bottom panel: evacuation with the RL-trained rescuer-leader. The exponent of the ``pseudo-gravitational'' potential \(\alpha = 3\). {Results are presented for 5000 runs with different random seeds.}} 
    \label{diagram_q}
\end{figure}

It is also worth to assess the evacuation effectivity. It may be measured by a number (or percentage) of evacuated individuals. However, if not all people are evacuated in a reasonable time, one cannot call the strategy effective and acceptable. Therefore, we decided to quantify it by another value -- by the probability,  that not all individuals are evacuated till time \(t\). The time dependence of this quantity is illustrated in \Cref{diagram_q}. Here we compare the probability that not all individuals are  evacuated  at time \(t\) for the case of evacuation by a trained rescuer-robot,  and that for the case of no leader, when the individuals move randomly in the dimly lit room, and randomly arrive to the exit zone; the robot is trained with the RL with the pseudo-gravitational encoding.  As it may be seen from the figure -- bottom panel, the RL-trained leader rescues the very last person at the time, when only about 50\%  individuals (worst scenario) are saved  for the case of no leader, upper panel (more precisely, the probability that evacuation is not complete is 50\%). \Cref{diagram_q} illustrates the dramatic increase of the number of saved people, when evacuation is guided by a trained leader. Here we demonstrate this for the most efficient strategy, associated with the pseudo-gravitational encoding  {(the network was trained once and statistics was collected for 5000 runs)}. Other RL methods would result in less effective evacuation. 

\section{Discussion and conclusion}

The problem of effective control of large ensembles of active particles is explored. We employ the approach where manipulation of such an ensemble is performed by a leader or group of leaders, within a leader-follower concept. As a result the problem reduces to the one for the effective strategy of the leader. For  the case of large uncertainty, or a noisy environment an application of reinforcement learning (RL) is the most appropriate. Here we combine RL with the method of artificial ”pseudo-gravitational” forces acting on the system. This significantly enhances the computational efficiency of RL. To describe the
guidance of a system of active particles by a leader we propose a generalized Vicsek
model; in this model active particles interact between each other and with the leader.  

The new method is then demonstrated in application to evacuation problem, when a crowd is to be rescued by a leader. That is, when a large group of people is to be escorted, as fast as possible, by the robot-rescuer (leader), from a  hazardous place towards an exit. 

{We show that although a straightforward RL with a basic feed-forward architecture requires slightly less (than ``pseudo-gravitational'') computational time per training step, it fails to converge to an efficient strategy. In practice, the obtained policy remains unstable and ineffective, despite the apparent speed advantage of the architecture.}
Even the most modern algorithm, based on the  transformer architecture with the RPO optimizer, yields rather poor efficiency. 
In contrast, an application of auxiliary ``pseudo-gravitational'' forces, acting between the leader and individuals (or groups of individuals) and between the leader and the exit, ultimately results in a very effective evacuation and {faster convergence}. These forces, defined as gradients of potentials, help the leader to estimate the optimal sequence of actions that is, the best strategy.

{An important advantage of our approach is that the pseudo-gravitational encoding is strictly independent of the number of individuals: the state is reduced to fixed-dimension collective features, so the same model can, in principle, be applied to any group size. By contrast, the feed-forward network requires retraining for each 
\(N\). The transformer, if properly configured (e.g., with pooling to a fixed dimension instead of concatenation and without fixed positional encodings), could also be independent of \(N\). Nevertheless, achieving true universality in practice requires a dedicated training protocol that explicitly includes episodes with varying group sizes — which is beyond the scope of the present study.}

The RL approach has been realized on the base of ANN with the application of the standard RPO algorithm and  the according software.  The evacuation efficiency is quantified,  in our study, by the probability of a complete evacuation till time \(t\). We observe that the training of the leader takes about \(6 \cdot 10^4\) time steps, which is relatively fast. We perform simulations to compare the evacuation effectivity for the case with a trained leader and that without a  leader. Our simulations demonstrate that the action of the trained rescuer dramatically increases the evacuation effectivity: at the time when all individuals are saved for the former case, only half (worst scenario) is rescued for the latter one (more precisely, the probability of complete evacuation is only 50\%).

As a proof of a concept, we consider here the space of  a simplest geometry -- a  square, and explore  an  action of a single robot-rescuer. In further studies we plan to explore more complicated geometry, with inner walls, multiple exits and  a group of robots-rescuers, acting synchronically. We expect,  that the application there of the combined model, comprising active-matter concepts with methods of artificial intelligence, like the RL, would be also effective. Moreover, we expect that the concept of ``pseudo-gravitational'' potentials, possibly, with the necessary generalization, would be fruitful for such studies as well. 

We believe that the integration of our ``pseudo-gravitational'' encoder into the neural architecture search (NAS) frameworks \citep{ren2021comprehensive,cheng2020hierarchical}  (as well as ZeroNAS \citep{yan2021zeronas}, and DS-Net++ \citep{li2022ds}) is a promising direction for  the future studies of an effective evacuation. 
 Furthermore, the Umbrella RL \citep{Nuzhin_2024_CNSNS}  also has great potential in this context.  
 Finally, the multi-agent version of the evacuation problem, with the MARLib \citep{hu2023marllib} framework, can be  considered -- it 
 would provide a powerful tool by incorporating a wide range of multi-agent RL algorithms and ANN architectures. Utilizing NAS frameworks, however,   would be computationally very costly, which remains a serious challenge for future studies.

\section*{Acknowledgments}

The work was supported by the Russian Science Foundation grant № 25-71-30008, \url{https://rscf.ru/project/25-71-30008/}. The authors also acknowledge the usage of the supercomputer  Zhores \citep{zhoresSuper}.

\appendix

\section{RL for a leader-guided evacuation.} \label{rl_overview}

Application of RL to active matter seems to be very fruitful, see e.g. \citep{Egor2021}. Here we present definitions and brief outline of the RL, applied for the generalized Vicsek model, that is, for an effective leader-guided evacuation of people from a hazardous place.

We are looking for an optimal strategy of a robot-rescuer with an aim to save as many people as possible, and as fast as possible. This  is 
a long-run strategy, based on the knowledge of the environment dynamics. We utilize a neural network, which may learn to find the trade-off between fast savings of the closest to an exit small groups and distant but large groups of people. Such a trade-off is modeled, formally, by  a reward function \(R_t\) from the reinforcement learning (RL) concept. In the RL one needs to maximize the expected reward for the whole series of successive actions. Moreover, the reward from instantaneous action is large, and becomes small for remote future actions. This is reflected in  - 
the {discounted sum of rewards} 
(also called "return"): 

\begin{equation}
\label{Rdef}
   R_{0} + \gamma R_{1} + \dots + \gamma^{t} R_{t} + \dots = \sum \limits_{t=0}^{\infty} \gamma^{t} R_t,
\end{equation}
which expected value is to be maximized. Here the discount factor \(\gamma \in \left(0; 1\right]\) indicates the preference of the immediate reward as compared 
to the delayed one.

The above equation helps us to formulate the rules, which the leader (robot-rescuer) should follow,  to achieve the goal. This is to be formalized in terms of required actions, undertaken in particular situations, and reward received for these actions. Then, one can apply methods of RL, which yield the strategy, leading to the maximal return.

The action policy \(\pi\) is the core function of the RL, which is seeking for an optimal policy -- optimal in terms of some performance measure (see below). The performance in the RL is assessed by  an expected {return} \Cref{Rdef}, discussed earlier. It is defined as  

\begin{equation}
\label{J}
   J(\pi) = \E_{\tau \sim \pi} \left[\sum_{t = 0}^\infty \gamma^t  R_t \right], 
\end{equation}
 Here \(\tau = (s_0,a_0,s_1,a_1,...)\) is a sequence of states \(s_t\) and actions \(a_t\) (at time \(t\)) called trajectory and \(R_t\) is the leader reward received in time \(t\). From this equation, one can see, that it is defined as 
 {the expected discounted sum of rewards, also known as the expected return}
 which the agent receives on its state-action trajectory. The optimal policy \(\pi^*\) is defined as a policy, that results in a maximal {expected return}, 

\begin{equation}
\label{pi_star}
   \pi^* = \arg \max_{\pi} J(\pi). 
\end{equation}
For a Markovian process the reward is determined by the function \(R_t = R(s_t,a_t,s_{t+1})\) -- the reward the agent receives for an action \(a_t\), made in a state \(s_t\) with the subsequent transition to the state \(s_{t+1}\). 
 
The action policy \(\pi (a|s) \) is optimized via stochastic gradient ascent, by a policy gradient algorithm, see e.g.  \citep{RL2}. Below we briefly sketch this approach, illustrating it for  the Actor-Critic algorithm, which is simpler to follow. We assume,  that the action policy corresponds to an ANN, specified by the  learnable parameters \(\theta\). Then, a sequence of episodes of the system evolution  is simulated  and the policy parameters are updated as 
\begin{align}
\label{weights_update}
\theta_{k+1}=\theta_{k}&+\alpha \nabla_{\theta} \hat{J}\left(\pi\right)\Big|_{{\pi = \pi_{\theta_{k}}}},
\nonumber
\end{align}
where \(\theta_{k}\) are ANN parameters at k-th episode, \(\alpha\) is a positive constant, known as the learning rate and 

\begin{equation}
     \nabla_{\theta} \hat{J}\left(\pi_\theta\right) =  \sum_{t=0}^{T} \nabla_{\theta} \log \pi_{\theta}\left(a_{t} \mid s_{t}\right) A\left(s_{t}, a_{t}\right)
\end{equation}
is the stochastic estimate of the gradient of the expected return,  \(J\), defined by \Cref{J}.
In the above equation \(A\) is the advantage function estimate \citep{schulman2015high}. It is calculated from the reward received during an episode simulation.
The advantage function serves to damp the gradient variance in the stochastic policy gradient estimate. This function, in its simplest implementation, reads, 
\begin{equation}
\label{A}
    A\left(s_{t}, a_{t}\right) = \sum \limits_{t'=t}^{T} \gamma^{t'} R_{t'} - V(s_t).
\end{equation}
where \(T\) is the episode duration\footnote{Note, that the formal definition implies the summation over infinite time, but for numerical reasons one can approximate it with a finite number of time steps, bounded by \(T\)} {and $V(s)$ is the state-value function, defined as the expected return obtained by the following policy $\pi$ starting from a state $s$:
\begin{equation}
V = \E_{\tau \sim \pi} \left[ \sum_{t=0}^{\infty} \gamma^t R_t \,\middle|\, s_0 = s \right].
\end{equation}
} The first term in the right-hand side (r.h.s.) in \Cref{A} is obtained in direct simulations of the environment with a current learnable policy \(\pi\).
From the definition of the value function \(V(s)\), e.g. \citep{RL2}, it 
follows,  that it may be estimated as the  minimum 
 of the expression, 
\begin{equation}
\label{value_loss}
    L(\hat{V}) = \E_{\tau \sim \pi} \left[\left(\sum \limits_{t'=t}^{T} \gamma^{t'} R_{t'} - \hat{V}(s_t)\right)^2\right].
\end{equation}
It is convenient to approximate \(V(s)\) by an ANN, using the parametrization with the (vectorial) parameter \(\eta\). Then the estimate for the value function is \(V_{\eta^*}(s)\), where

\begin{equation}
    \eta^* = \arg \min_{\eta}L(V_\eta).
\end{equation}
The external parameter \(\eta^*\) may be found by the standard stochastic gradient approach, which updates \(\eta\) in the successive episodes: 
\begin{align}
\label{weights_update_v}
\eta_{k+1}= \eta_{k} -  \beta  \nabla_{\eta}\hat{L}(V_\eta)\Big|_{V_\eta = V_{\eta_k}},
\end{align}
where \(\eta_{k}\) are ANN parameters at k-th episode, \(\beta\) is a positive constant -- the learning rate {and 
\begin{equation}
    \hat{L}(\hat{V}) = \left(\sum \limits_{t'=t}^{T} \gamma^{t'} R_{t'} - \hat{V}(s_t)\right)^2
\end{equation}
is stochastic estimate of the loss function of the value function,  \Cref{value_loss}}.

\begin{algorithm*}[ht]\small
        \textbf{Def.:} \(\pi\) --- policy, \(v\) --- value, \texttt{env} --- environment
    \begin{algorithmic}[1]
    \FOR{\textsc{each episode} \texttt{ep}  \textsc{of training}}  
            \STATE {
                \texttt{observation, \_ = env.reset()}
                }
            \STATE {
                \texttt{gravity\_observation = gravity\_encoding(observation)}
                }
            \COMMENT{application of pseudo-gravitational encoding}
            \FOR{\textsc{each step} \texttt{t}  \textsc{of the episode}}
                    \STATE {
                        \texttt{action = policy(gravity\_observation)}
                        }
                    \STATE {
                        \texttt{observation, reward, terminated, truncated, \_ = env.step(action)}
                        }
                    \STATE {
                        \texttt{\text{gravity\_observation} = \texttt{gravity\_encoding(observation)}}
                        }
                    \COMMENT{application of pseudo-gravitational encoding}
            \ENDFOR
            \STATE \texttt{policy update with PPO,RPO, A2C, etc.}
            \COMMENT{based on gravity\_observation and received reward}
    \ENDFOR
    \end{algorithmic}
    \caption{Learning with Pseudo-Gravitational encoding}
    \label{alg:learninig}
    \end{algorithm*}

The gradient is calculated using automatic differentiation packages; for more detail see the original paper  \citep{mnih2016asynchronous}, where A2C is discussed.
In practice, however, we utilize a more powerful  RPO  algorithm \citep{rahman2022robust} that is a robust version of PPO \citep{schulman2017proximal} algorithm. The main idea of PPO coincides with that of A2C, but it contains some parts,
which significantly improve the speed and convergence. 
Currently, the PPO is a widely spread, state-of-the-art  RL algorithm.  Schematically, the training with pseudo-gravitational encoding, is present in the \Cref{alg:learninig}.

\section{Average angle/direction}
\label{sec:average_angle}

The dynamic rules for the Vicsek  model require 
computation of the average direction of the  moving neighbouring individuals. Here one needs to apply correct 
summing rules for  the angles. For instance,  for two particles, the expression for the average angle reads \citep{labasken}:
\begin{equation}
    \text{MEAN}\left(\theta_{1}, \theta_{2}\right) =
    \begin{cases}
    \frac{\theta_{1}+\theta_{2}}{2} & \text { if } \quad\left|\theta_{1}-\theta_{2}\right| \leq \pi, \\ 
    \frac{\theta_{1}+\theta_{2}}{2}+\pi & \text { if } \quad\left|\theta_{1}-\theta_{2}\right|>\pi .
    \end{cases}
\end{equation}

For \(N\) particles, the according expression becomes more complicated; it is based on the trigonometric relation. First, we define the ratio: 
\begin{equation}
\label{eq:p_xy}
    p = \frac{y_p}{x_p} = \frac{\sin{\theta_1} + \hdots+\sin{\theta_N}}{\cos{\theta_1} + \hdots+\cos{\theta_N}}, 
\end{equation}
then the average angle is defined as follows\footnote{The relations below correspond to the standard function \texttt{atan2(\(y_p,x_p\))}.}
\begin{equation} \label{anglesystem}
    \text{mean}\left(\theta\right) =
    \begin{cases}
    \arctan{p} & \text { if } \quad x_p > 0, \\ 
    \arctan{p} + \pi & \text { if } \quad x_p < 0 \wedge y_p \geq 0, \\
    \arctan{p} - \pi & \text { if } \quad x_p < 0 \wedge y_p < 0, \\
    \pi/2 & \text { if } \quad x_p = 0 \wedge y_p > 0, \\
    -\pi/2 & \text { if } \quad x_p = 0 \wedge y_p < 0, \\ 
    \texttt{undefined} & \text { if } \quad x_p = 0 \wedge y_p = 0.
    \end{cases}
\end{equation}

{
For the case of two individuals with angles $\theta_1, \theta_2$ and positive weights $w_1, w_2$, 
Eq. \eqref{eq:p_xy} is modified to
\begin{equation}
\label{eq:p_xy_weigted}
p = \frac{y_p}{x_p} = \frac{w_1 \sin \theta_1 + w_2 \sin \theta_2}{\,w_1 \cos \theta_1 + w_2 \cos \theta_2}.
\end{equation}
The corresponding mean angle is then defined exactly as in Eq. \eqref{anglesystem}, applied to these weighted sums. 
Such weighted averaging is used in the update rule of Eq. (\ref{MVM_update_rule}) in the main text.
}

{Note that Eqs. \eqref{anglesystem} and \eqref{eq:p_xy_weigted} are equivalent to the averaging of unit direction vectors and then taking the argument of the resulting vector, which in mathematical  notations reads
\begin{equation}
\text{mean}\left(\theta\right) = \arg\!\Bigl[w_1 e^{i\theta_1} + w_2 e^{i\theta_2}\Bigr]
\end{equation}
and corresponds to the standard formulation of the Vicsek model~\cite{VicsekINITIAL}. 
We use the angle notation here for compactness only. 
It is important to stress that a simple arithmetic mean of angles is not 
well-defined, since its result depends on the chosen branch of the angle 
interval and may yield contradictory outcomes. 
By contrast, averaging in the vector form and then computing the argument 
ensures a unique definition of the mean direction. 
In degenerate symmetric cases (when the average vector has zero magnitude), 
no prevailing orientation exists and the new direction is undefined which is also reflected in the equation.}

\section{Experimental results}
\label{apx:exp_res}

\begin{figure*}[ht!]
    \centering
     {\includegraphics[width =0.49\textwidth]{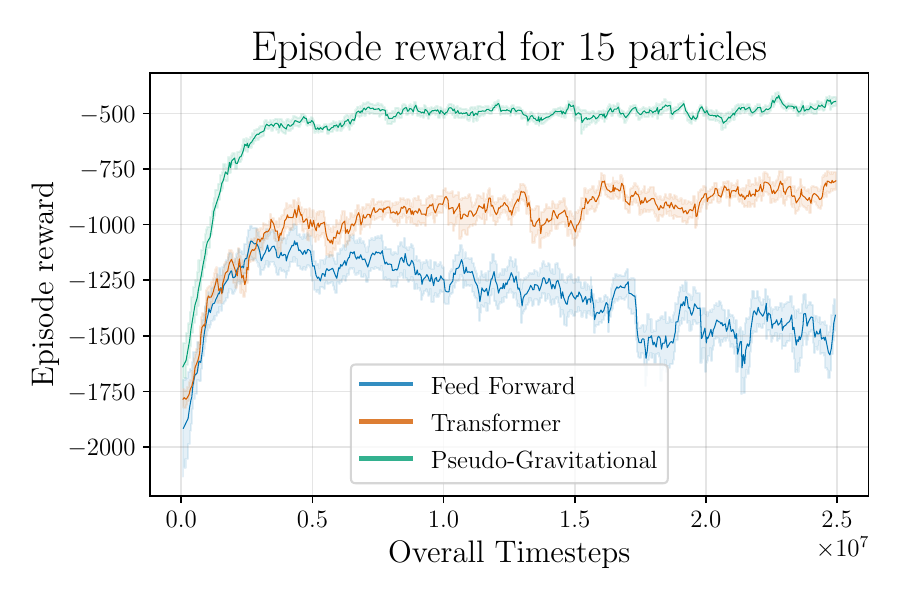}}
     \hfil
    {\includegraphics[width =0.49\textwidth]{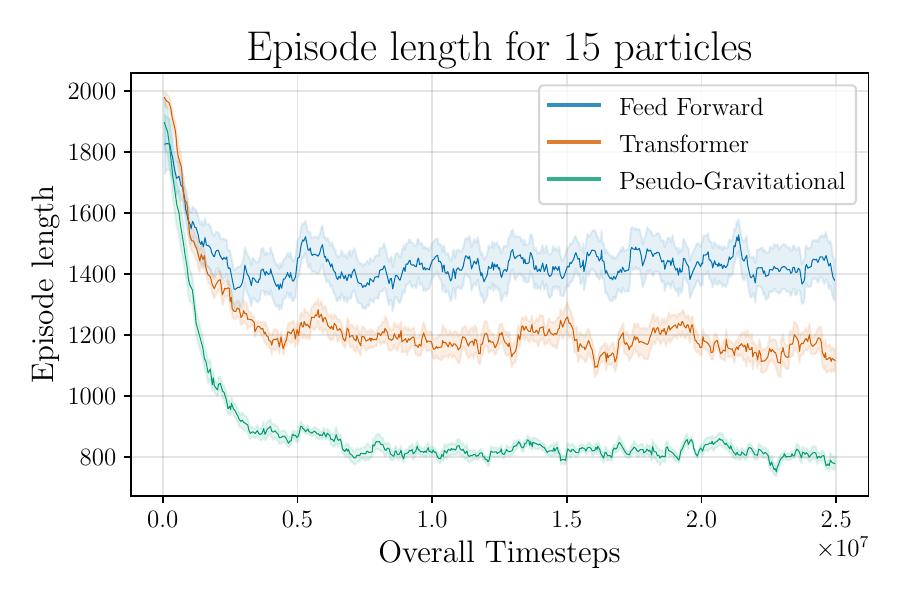}}
    \hfil
    {\includegraphics[width =0.49\textwidth]{figures/reward-exp_name-60.pdf}}
    \hfil
    {\includegraphics[width =0.49\textwidth]{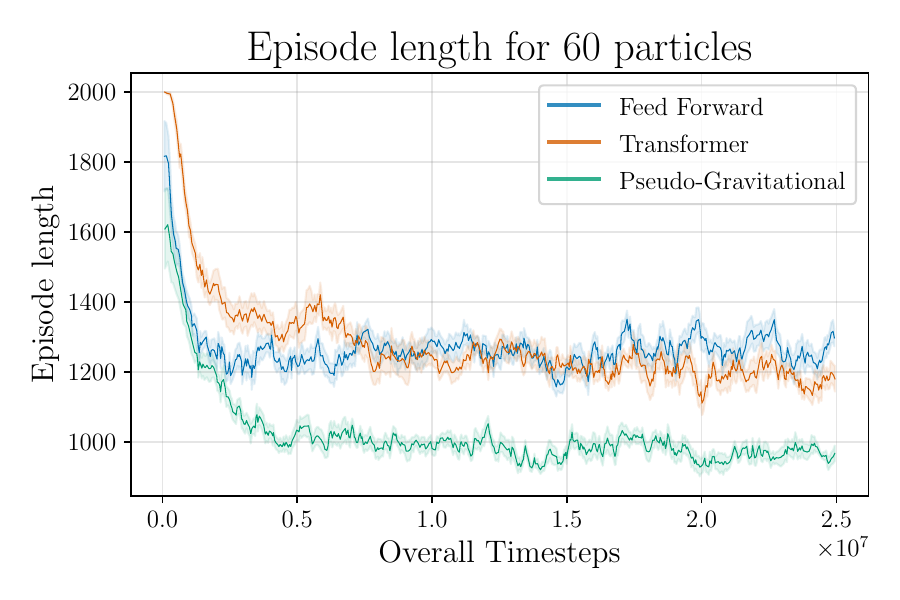}}
    
    \caption{Training curves for various values 60 and 15 individuals. Average episode length and episode reward are depicted. The results are average in 10 training runs {with different random seeds}; the shaded areas depict the standard error.}
    \label{alpha_evac1}
\end{figure*}

\Cref{alpha_evac1} presents average episode length and episode reward training curves for 60 and 15 individuals for 3 method: Feed Forward, Transformer and Pseudo-Gravitational. More details on methods in \Cref{subsec:methods}. The average duration of training 25M timesteps for each method is shown in \Cref{num:table:Duration}. Note that Pseudo-Gravitational approach reaches optimal value at $\sim$3M timesteps.
 
\begin{table}[ht!]
\caption{Duration of training for \Cref{diagram_training,alpha_evac1} (25M timesteps)}
\label{num:table:Duration}
\begin{center}
    \begin{tabular}{ | p{5cm} | r | r |}
        \hline \hline
        Method & 15 particles & 60 particles  \\ \hline \hline
        Feed Forward & 8h 54m & 9h 16m \\ \hline
        Transformer & 14h 58m & 1d 9h 25m \\ \hline
        Pseudo-Gravitational & 9h 27m & 10h 5m  \\ \hline
    \end{tabular}
\end{center}
\end{table}

\begin{figure*}[ht!]
    \centering
     {\includegraphics[width =0.49\textwidth]{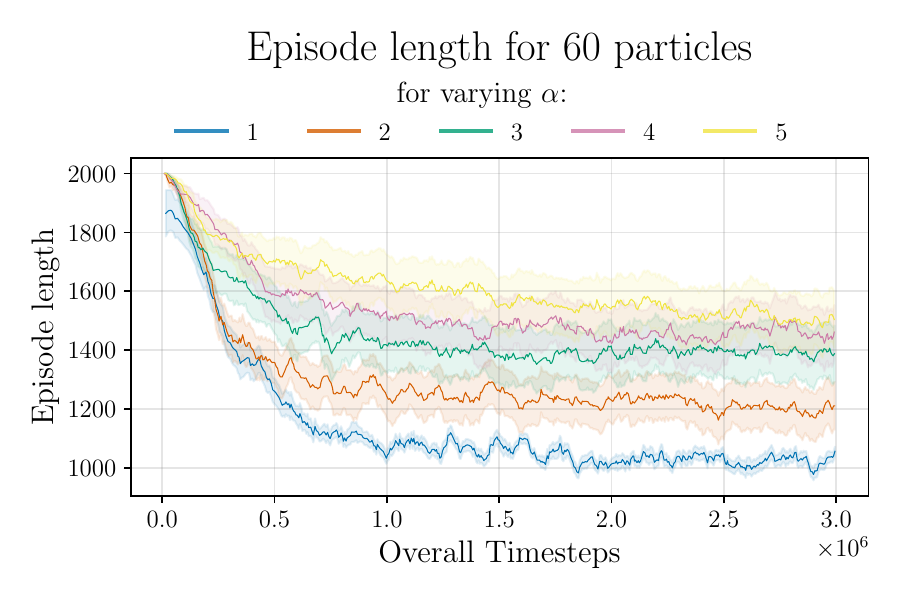}} 
     \hfil
    {\includegraphics[width =0.49\textwidth]{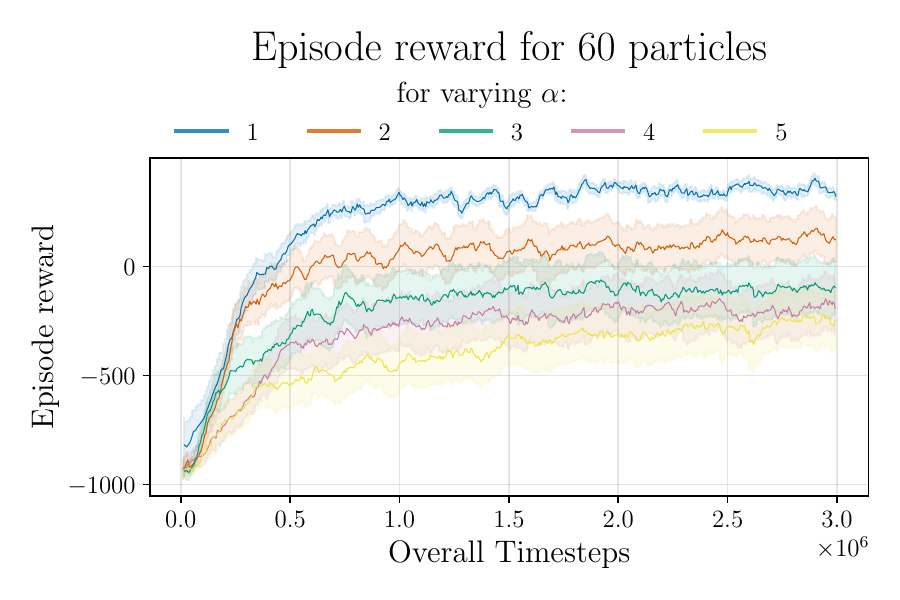}} 
    
    \centering
     {\includegraphics[width =0.49\textwidth]{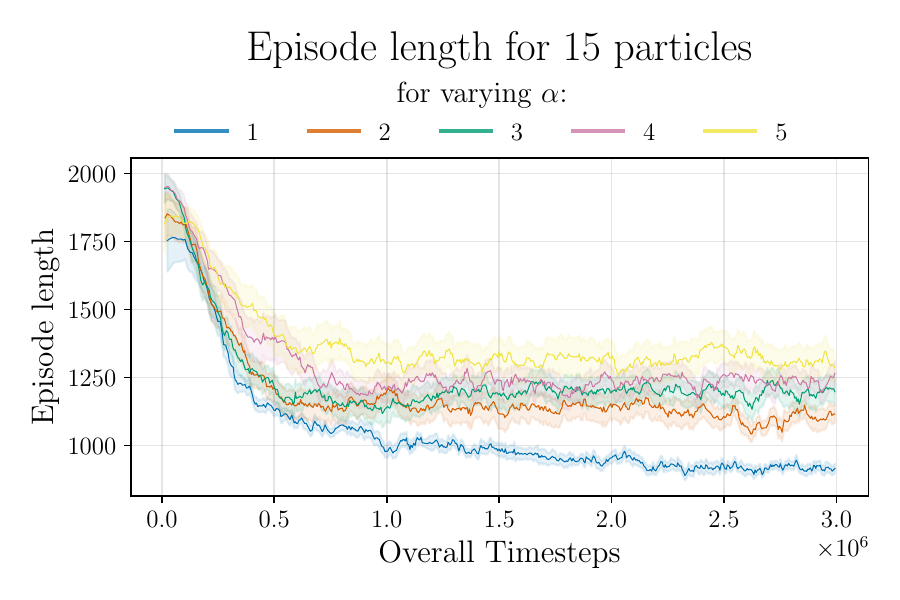}} 
     \hfil
    {\includegraphics[width =0.49\textwidth]{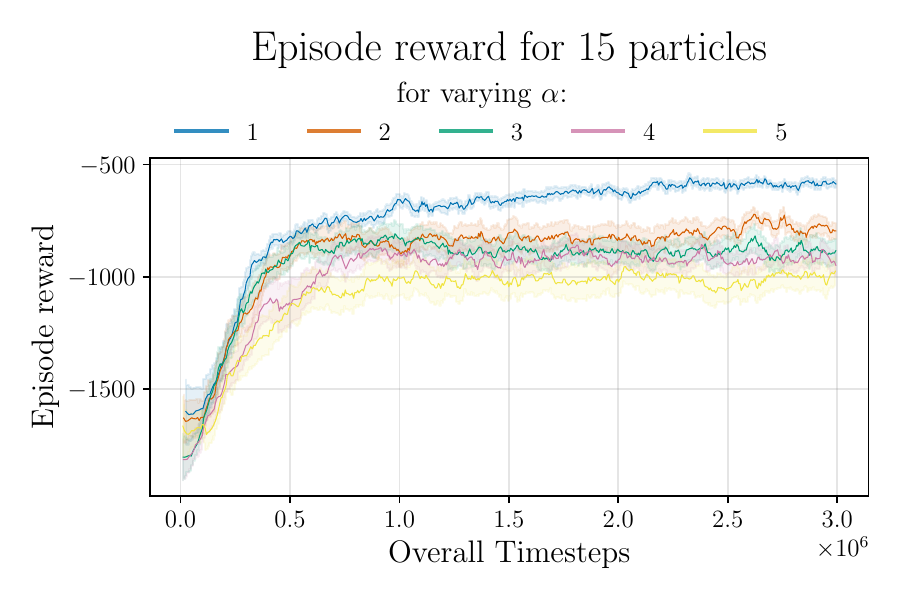}} 
    
        \caption{Training curves for various values of the exponent \(\alpha\)  for the RL with the pseudo-gravitational encoding. Average episode length and episode reward are depicted. The results are average in 20 seeds; the shaded areas depict the standard error.}
        \label{alpha_evac}
\end{figure*}

\newpage
\section{Parameters of the model and ``Pseudo-Gravitational'' potential}

\Cref{num:table:Vicsek}  shows all the default parameters affecting the experiment. Such parameters for the environment were used in order to conduct realistic modeling of human behavior.

\begin{table}[ht]
\caption{Model parameters }
\label{num:table:Vicsek}
\begin{center}
    \begin{tabular}{ | p{2cm} | p{4cm} | c |}
        \hline \hline
        Parameter & Description & Value  \\ \hline \hline
        \(L\), units & Square simulation domain: \([-L, L]\) in both \(x\) and \(y\) directions & \(1\) \\ \hline 
        \(r\), units & Vicsek radius & \(0.1\) \\ \hline 
        \(\frac{r_{\rm L}}{r}\) & Relative leader's radius &  \(2\) \\ \hline
        \(\zeta\) & Noise intensity  & \(0.2\) \\ \hline
        \(v\), units/time step & Absolute velocity & \(0.01\) \\ \hline
        \(t_{max}\), time steps &  {Max episode length} & \(2000\) \\ \hline
        \(N\) & Number of individuals & \(60\) \\ \hline
        \(r_{exit}\) units & Radius of \emph{Exit Zone} & \(0.4\) \\ \hline
        {\(q\)} & {Enslaving degree} & \(1\) \\ 
        \hline
    \end{tabular}
\end{center}
\end{table}

The main parameter of the model is the Vicsek  radius. Varying the Vicsek radius corresponds to the variation of the visibility area for each individual -- the larger the radius, the greater  the visibility area.

An important training hyperparameter is the exponent \(\alpha\) of the pseudo-gravitational encoding [see the \Cref{gravity}]. \Cref{alpha_evac} demonstrates the  experiments to determine the most efficient \(\alpha\) in terms of optimization speed and performance. The figure suggests, that the optimal value is  \(\alpha\) = 1.

\begin{table}[ht]
\caption{Network parameters}
\label{num:table:Network}
\begin{center}
    \begin{tabular}{ | p{4cm} | p{2cm} | c |}
        \hline \hline
        Description & Parameter & Value  \\ \hline \hline
        Dropout Rate & dropout & 0.1 \\ \hline
        RPO Alpha & rpo\_alpha & 0.5 \\ \hline
        Hidden Dimension & num\_hidden & 64 \\ \hline
        Number of Layers & num\_layers & 3 \\ \hline
    \end{tabular}
\end{center}
\end{table}

\begin{table}[ht]
\caption{RPO agent parameters}
\label{num:table:RPOAgent}
\begin{center}
    \begin{tabular}{ | p{5cm} | p{4cm} | c |}
        \hline \hline
        Description & Parameter & Value  \\ \hline \hline
        Discount Factor & gamma & 0.99 \\ \hline
        Value Function Coefficient & vf\_coef & 0.5 \\ \hline
        Entropy Coefficient & ent\_coef & 0 \\ \hline
        Normalize Advantage & norm\_adv & true \\ \hline
        Number of Environments & num\_envs & 3 \\ \hline
        Anneal Learning Rate & anneal\_lr & true \\ \hline
        Clip Coefficient & clip\_coef & 0.2 \\ \hline
        {Rollout length}
        & num\_steps & 2048 \\ \hline
        Target KL Divergence & target\_kl & null \\ \hline
        Clip Value Loss & clip\_vloss & true \\ \hline
        GAE Lambda & gae\_lambda & 0.95 \\ \hline
        Learning Rate & learning\_rate & 0.0005 \\ \hline
        Max Gradient Norm & max\_grad\_norm & 0.5 \\ \hline
        Update Epochs & update\_epochs & 10 \\ \hline
        Number of Minibatches & num\_minibatches & 32 \\ \hline
        Total Timesteps & total\_timesteps & {25000000} \\ \hline
        Deterministic Mode & torch\_deterministic & true \\ \hline
    \end{tabular}
\end{center}
\end{table}

\begin{table}[ht]
\caption{Transformer parameters}
\label{num:table:TransformerBlock}
\begin{center}
    \begin{tabular}{ | p{5cm} | p{3cm} | c |}
        \hline \hline
        Description & Parameter & Value  \\ \hline \hline
        Number of Heads & num\_heads & 3 \\ \hline
        Use Residual Connections & use\_resid & false \\ \hline
        Number of Blocks & num\_blocks & 2 \\ \hline
        Feedforward Dimension & dim\_feedforward & 96 \\ \hline
        Model Dimension & d\_model & 6 \\ \hline
    \end{tabular}
\end{center}
\end{table}

\vskip 0.2in
\newpage
\clearpage

\bibliographystyle{elsarticle-num} 
\bibliography{general}
\end{document}